\documentclass[final]{cvpr}

\usepackage{times}
\usepackage{epsfig}
\usepackage{graphicx}
\usepackage{amsmath}
\usepackage{amssymb}
\usepackage{graphicx}
\usepackage{epstopdf}
\usepackage{makecell}
\usepackage{multirow}
\usepackage{array}
\usepackage{bm}
\usepackage{dsfont}
\usepackage{color}
\usepackage{caption}
\usepackage{algorithm} 
\usepackage{algorithmic} 
\usepackage{graphicx}
\usepackage{subfigure}
\usepackage{xcolor}

\usepackage{pifont}

\newcommand{\tabincell}[2]{\begin{tabular}{@{}#1@{}}#2\end{tabular}}

\usepackage{enumitem}
\newenvironment{tight_itemize}{
\begin{itemize}[leftmargin=20pt]
  \setlength{\topsep}{0pt}
  \setlength{\itemsep}{0pt}
  \setlength{\parskip}{0pt}
  \setlength{\parsep}{0pt}
}{\end{itemize}}

\usepackage[pagebackref=true,breaklinks=true,colorlinks,bookmarks=false]{hyperref}

\def\cvprPaperID{701} 
\def\confYear{CVPR 2021}

\begin{document}
\captionsetup{font=footnotesize}
\title{WebFace260M: A Benchmark Unveiling the Power of   Million-Scale \\ Deep Face Recognition}

\author{Zheng Zhu\textsuperscript{\rm 1,2*} ~ Guan Huang\textsuperscript{\rm 2\thanks{These authors contributed equally to this work.}} ~ Jiankang Deng\textsuperscript{\rm 3} ~Yun Ye\textsuperscript{\rm 2} ~Junjie Huang\textsuperscript{\rm 2} \\ ~Xinze Chen\textsuperscript{\rm 2}  ~Jiagang Zhu\textsuperscript{\rm 2} Tian Yang\textsuperscript{\rm 2} ~Jiwen Lu\textsuperscript{\rm 1} ~Dalong Du\textsuperscript{\rm 2} ~Jie Zhou\textsuperscript{\rm 1}\\
\textsuperscript{\rm 1}Tsinghua University
~ ~ \textsuperscript{\rm 2}XForwardAI
~ ~ \textsuperscript{\rm 3}Imperial College London\\
\tt\small \{zhengzhu,lujiwen\}@tsinghua.edu.cn    \{guan.huang,dalong.du\}@xforwardai.com \\
\tt\small j.deng16@imperial.ac.uk
}


\maketitle
\begin{abstract}
In this paper, we contribute a new million-scale face benchmark containing \textbf{noisy 4M identities/260M faces} (WebFace260M) and \textbf{cleaned 2M identities/42M faces} (WebFace42M) training data, as well as an elaborately designed time-constrained evaluation protocol. Firstly, we collect 4M name list and download 260M faces from the Internet. Then, a Cleaning Automatically utilizing Self-Training (CAST) pipeline is devised to purify the tremendous WebFace260M, which is efficient and scalable. To the best of our knowledge, the cleaned WebFace42M is the largest public face recognition training set and we expect to close the data gap between academia and industry. Referring to practical scenarios, Face Recognition Under Inference Time conStraint (FRUITS) protocol and a test set are constructed to comprehensively evaluate face matchers.

Equipped with this benchmark, we delve into million-scale face recognition problems. A distributed framework is developed to train face recognition models efficiently without tampering with the performance. Empowered by WebFace42M, we reduce relative 40\% failure rate on the challenging IJB-C set, and \textbf{ranks the 3rd among 430 entries on NIST-FRVT}. Even 10\% data (WebFace4M) shows superior performance compared with public training set. Furthermore, comprehensive baselines are established on our rich-attribute test set under FRUITS-100ms/500ms/1000ms protocol, including MobileNet, EfficientNet, AttentionNet, ResNet, SENet, ResNeXt and RegNet families.
Benchmark website is \url{https://www.face-benchmark.org}.
\end{abstract}
\vspace{-0.8cm}
\section{Introduction}
\vspace{-0.1cm}
Recognizing faces in the wild has achieved a remarkable success due to the boom of CNNs.
The key engine of recent face recognition consists of network architecture evolution~\cite{AlexNet,GoogleNet,VGGNet,ResNet,BN,Dropout,he2015delving,ShuffleNet,MobileNet,wu2018light,Deepid3}, a variety of loss functions~\cite{DeepFace,FaceNet,DeepID,DeepID2,wen2016discriminative,deng2017marginal,ranjan2018crystal,NormFace,liu2016large,A-SoftMax,CosFace,AM-SoftMax,ArcFace,Curricularface}, and growing face benchmarks~\cite{LFW,AgeDB,CFP,CALFW,CPLFW,MegaFace,IJB-A,IJB-B,IJB-C,CASIA-WebFace,VGGFace,VGGFace2,UMDFaces,MF2,MS1M,IMDB-Face}.

\begin{figure}[!t]
\centering
\includegraphics[width=1.0\linewidth]{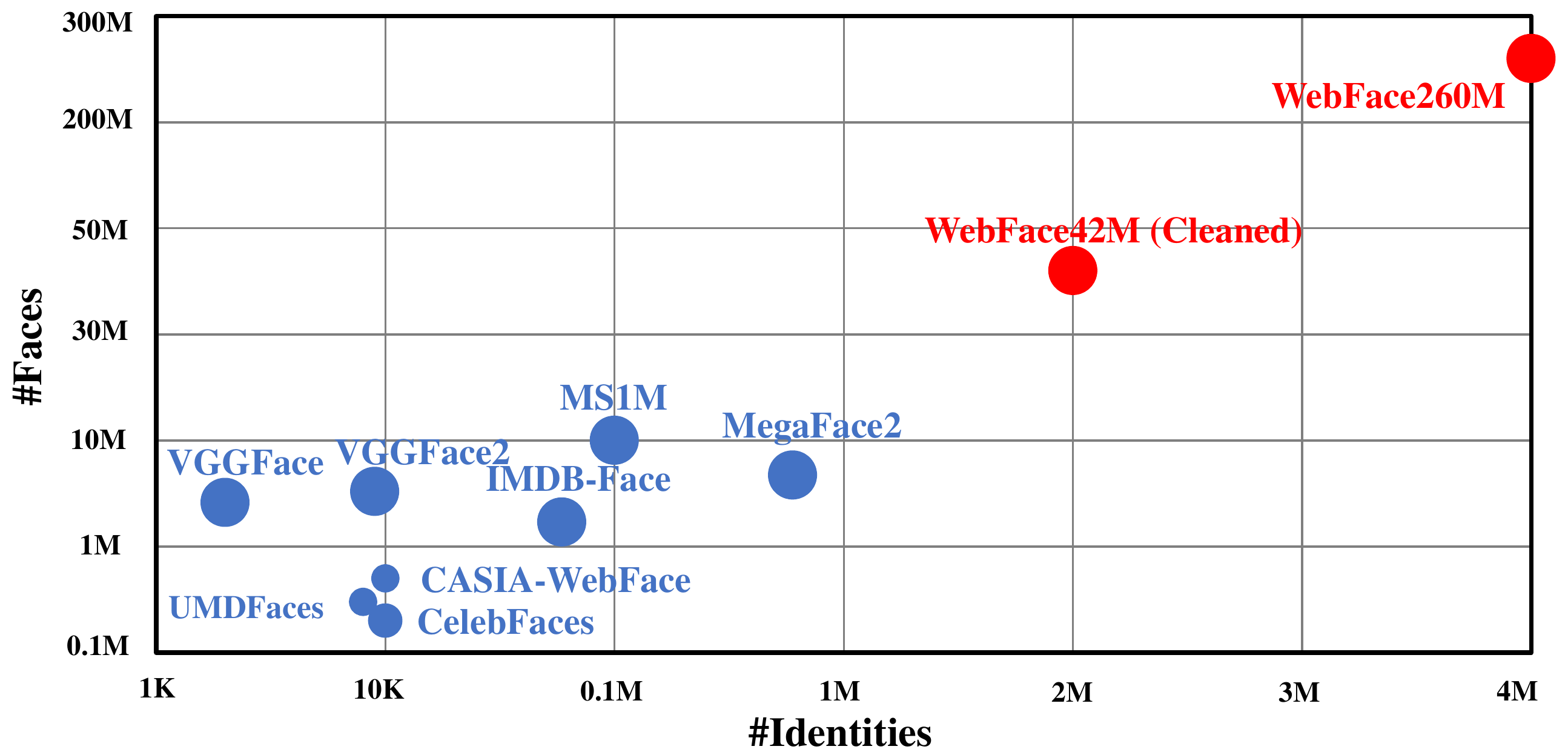}
\caption{Comparisons of \# identities and \# faces for our WebFace data and public training set.}
\label{fig:training_set}
\vspace{-4mm}
\end{figure}

\begin{table*}[!t]
\begin{center}{\scalebox{0.85}{
\begin{tabular}{l|c|c|c|c|c|c|c}
\hline
Dataset & \# Identities & \# Images & Images/ID  & Cleaning  & \# Attributes & Availability & Publications\\ \hline
\hline
CASIA-WebFace~\cite{CASIA-WebFace} & $10$ K & $0.5$ M & 47 & Auto & - & Public & Arxiv 2014\\
CelebFaces~\cite{DeepID} & $10$ K & $0.2$ M & 20 & Manual & 40  & Public & ICCV 2015\\
UMDFaces~\cite{UMDFaces} & $8$ K & $0.3$ M & 45 & Semi-auto & 4 & Public & IJCB 2017\\
VGGFace~\cite{VGGFace} & $2$ K & $2.6$ M & 1,000 & Semi-auto & - & Public & BMVC 2015\\
VGGFace2~\cite{VGGFace2} & $9$ K & $3.3$ M & 363 & Semi-auto & 11 & Public & FG 2018\\
MS1M~\cite{MS1M} &$0.1$ M & $10$ M & 100 & No & -& Public & ECCV 2016\\
MS1M-IBUG~\cite{deng2017marginal} &$85$ K & $3.8$ M & 45 & Semi-auto & -& Public & CVPRW 2017\\
MS1MV2~\cite{ArcFace} & $85$ K & $5.8$ M & 68 & Semi-auto & -& Public & CVPR 2019\\
MS1M-Glint~\cite{glintweb} & $87$ K & $3.9$ M  & 44 & Semi-auto & -& Public & -\\
MegaFace2~\cite{MF2} & $0.6$ M & $4.7$ M & 7 & Auto & -& Public & CVPR 2017\\
IMDB-Face~\cite{IMDB-Face} & $59$ K & $1.7$ M & 29 & Manual &  - & Public & ECCV 2018\\ \hline
\textcolor[RGB]{128,128,128}{Facebook~\cite{DeepFace}} & \textcolor[RGB]{128,128,128}{$4$ K}  & \textcolor[RGB]{128,128,128}{$4.4$ M} & \textcolor[RGB]{128,128,128}{1,100} & \textcolor[RGB]{128,128,128}{-} & \textcolor[RGB]{128,128,128}{-}& \textcolor[RGB]{128,128,128}{Private} &\textcolor[RGB]{128,128,128}{CVPR 2014}\\
\textcolor[RGB]{128,128,128}{Facebook~\cite{taigman2015web}} & \textcolor[RGB]{128,128,128}{$10$ M} & \textcolor[RGB]{128,128,128}{$500$ M} & \textcolor[RGB]{128,128,128}{50} & \textcolor[RGB]{128,128,128}{-} & \textcolor[RGB]{128,128,128}{-}& \textcolor[RGB]{128,128,128}{Private} &\textcolor[RGB]{128,128,128}{CVPR 2015}\\
\textcolor[RGB]{128,128,128}{Google~\cite{FaceNet}} & \textcolor[RGB]{128,128,128}{$8$ M} & \textcolor[RGB]{128,128,128}{$200$ M} & \textcolor[RGB]{128,128,128}{25} & \textcolor[RGB]{128,128,128}{-} & \textcolor[RGB]{128,128,128}{-}& \textcolor[RGB]{128,128,128}{Private} & \textcolor[RGB]{128,128,128}{CVPR 2015}\\
\textcolor[RGB]{128,128,128}{MillionCelebs~\cite{MillionCelebs}} & \textcolor[RGB]{128,128,128}{$0.6$ M} & \textcolor[RGB]{128,128,128}{$18.8$ M} & \textcolor[RGB]{128,128,128}{30} & \textcolor[RGB]{128,128,128}{Auto} & \textcolor[RGB]{128,128,128}{-}& \textcolor[RGB]{128,128,128}{Private} & \textcolor[RGB]{128,128,128}{CVPR 2020}\\ \hline
\textbf{WebFace260M}  & \textbf{4 M}  & \textbf{260M}  & \textbf{65} & No & -& Public & - \\
\textbf{WebFace42M} & \textbf{2 M}  & \textbf{42M}  & \textbf{21} & Auto & 7 & Public & - \\ \hline
\end{tabular}}}
\end{center}
\vspace{-4mm}
\caption{Training data for deep face recognition. The cleaned WebFace42M is the largest public training set in terms of both \# identities and \# images.}
\vspace{-6mm}
\label{table:training_set}
\end{table*}

Face benchmarks empower researchers to train and evaluate high-performance face recognition systems. Even though growing efforts have been devoted to investigating sophisticated networks \cite{chen2018mobilefacenets,yan2019vargfacenet,VGGFace2,wang2020hierarchical,fanggenerate} and losses \cite{A-SoftMax,CosFace,AM-SoftMax,ArcFace,Curricularface,circleloss,subcenter}, academia is restricted by limited training set and nearly saturated test protocols. As shown in Tab.\ref{table:training_set}, the public largest training sets in terms of identities and faces are MegaFace2~\cite{MF2} and MS1M~\cite{MS1M}, respectively. MegaFace2 contains 4.7M faces of 672K subjects collected from Flickr~\cite{Flickr}. MS1M consists of 10M faces of 100K celebrities but the noise rate is around 50\%~\cite{IMDB-Face}. In contrast, companies from industry can access much larger private data to train face recognition models: Google utilizes 200M images of 8M identities to train FaceNet~\cite{FaceNet}, and Facebook~\cite{taigman2015web} performs training by 500M faces of 10M identities. This data gap hinders researchers to push the frontiers of deep face recognition. Main obstacles for tremendous training data lie in large-scale identity collection, effective and scalable cleaning, and efficient training.

On the other hand, evaluation protocols and test set play an essential role in analysing face recognition performance.
Popular evaluations for face recognition including LFW families~\cite{LFW,CALFW,CPLFW}, CFP~\cite{CFP}, AgeDB~\cite{AgeDB}, RFW~\cite{RFW}, MegaFace~\cite{MegaFace}, IJB families~\cite{IJB-A,IJB-B,IJB-C} mainly target the pursuit of the accuracy, which have been almost saturated recently. In real-world application scenarios, face recognition is always restricted by the inference time, such as unlocking mobile telephone with smooth experience. Lightweight face recognition challenge~\cite{LFR} takes a step toward this goal, but it neglects the time cost of detection and alignment. To the best of our knowledge, NIST-FRVT~\cite{FRVT} is the only time-constrained face recognition protocol. However, strict submission policy (no more than one submission every four calendar months) hinders researchers to freely evaluate their algorithms.

To address the above problems, this paper constructs a new large-scale face benchmark consists of \textbf{4M identities/260M faces} (WebFace260M) as well as a time-constrained assessment protocol. Firstly, a name list of 4M celebrities is collected and 260M images are downloaded utilizing a search engine. Then, we perform Cleaning Automatically by Self-Training (CAST) pipeline, which is scalable and does not need any human intervention. The proposed CAST procedure results in \textbf{high-quality 2M identities and 42M faces} (WebFace42M). With such data size, a distributed training framework is developed to perform efficient optimization.
Referring to various real-world applications, we design the Face Recognition Under Inference Time conStraint (FRUITS) protocol, which enables academia to evaluate deep face matchers comprehensively. The FRUITS protocol consists of 3 tracks: 100, 500 and 1000 milliseconds. Since public evaluations are most saturated ~\cite{LFW, AgeDB, CFP} and may contain noise~\cite{MegaFace,IJB-C}, we manually construct a new test set with rich attributes to enable FRUITS, including different age, gender, race and scenario evaluations. This test set will be actively maintained and updated.


Based on the proposed new large-scale benchmark, we delve into million-scale deep face recognition problems. The distributed training approach could be performed at near linear acceleration without performance drops. Verification accuracy on public dataset indicates that the proposed million-scale training data is indispensable to push the frontiers of deep face recognition: WebFace42M achieves 97.70\% TAR@FAR=1e-4 on challenging IJB-C \cite{IJB-C} under standard ResNet-100 configurations, relatively reducing near 40\% error rate compared with public state-of-the-arts. 10\% of our data (WebFace4M) also obtains superior performance than similar-sized MS1M families \cite{deng2017marginal,ArcFace,glintweb} and MegaFace2 \cite{MF2}. Furthermore, we participate in the \textbf{NIST-FRVT}~\cite{FRVT} and \textbf{ranks the 3rd among 430 entries} based on WebFace42M.
Finally, comprehensive face recognition systems are evaluated under FRUITS-100ms/500ms/1000ms protocols, including MobileNet \cite{MobileNet,chen2018mobilefacenets}, EfficientNet \cite{tan2019efficientnet}, AttentionNet \cite{AttentionNet}, ResNet \cite{ResNet}, SENet \cite{SENet}, ResNeXt \cite{ResNeXt} and RegNet families \cite{RegNet}.
With this new face benchmark, we hope to close the data gap between the research community and industry, and facilitate the time-constrained recognition performance assessment for real-world applications.

The main contributions can be summarized as follows:
\vspace{-0.2cm}
\begin{tight_itemize}
\item A large-scale face recognition dataset is constructed for the research community towards closing the data gap behind the industry.
The proposed WebFace260M consists of 4M identities and 260M faces, which provides an excellent resource for million-class deep face cleaning and recognition as shown in Fig.\ref{fig:training_set} and Tab.\ref{table:training_set}.
\item We contribute the largest training set WebFace42M which sets new SOTA on challenging IJB-C and ranks the 3rd on NIST-FRVT. This cleaned data is automatically purified from WebFace260M by a scalable and effective self-training pipeline.
\item The FRUITS protocol as well as a test set with rich attributes are constructed to facilitate the evaluation of real-world applications. A series of tracks are designed referring to different deployment scenarios.
\item Based on the new benchmark, we perform extensive million-scale face recognition experiments. Enabled by distributed training framework, comprehensive baselines are established on our test set under the FRUITS protocol. The results indicate substantial improvement room for light-weight track, as well as the necessity of innovation in heavy-weight track.
\end{tight_itemize}

\section{WebFace260M and WebFace42M}

\noindent{\bf Celebrity name list and image collection.}
Knowledge graphs website Freebase~\cite{freebase} and well-curated website IMDB~\cite{imdb_website} provide excellent resources for collecting celebrity names. Furthermore, commercial search engines such as Google~\cite{google_image} make it possible to collect images of a specific identity with ranked correlation. Our celebrity name list consists of two parts: the first one is borrowed from MS1M (1M, constructed from Freebase) and the second one is collected from the IMDB database. There are nearly 4M celebrity names in the IMDB website, while we found some subjects have no public image from search engines. Therefore, only 3M celebrity names in IMDB are chosen for our benchmark. Based on the name list, celebrity faces are searched and downloaded via Google image search engine. 200 images per identity are downloaded for top 10\% subjects, while 100, 50, 25 images are reserved for remaining 20\%, 30\%, 40\% subjects, respectively. Finally, we collect 4M identities and 265M images.

\begin{figure}
\small
\centering
\subfigure[Date of Birth]{
\label{fig:allbirth}
\includegraphics[width=0.13\textwidth]{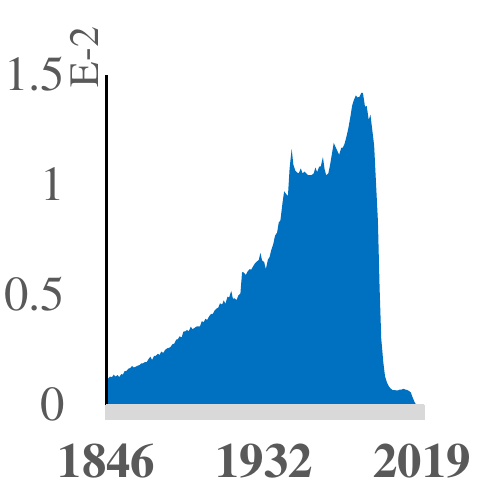}}
\subfigure[Nationality]{
\label{fig:allnationality}
\includegraphics[width=0.14\textwidth]{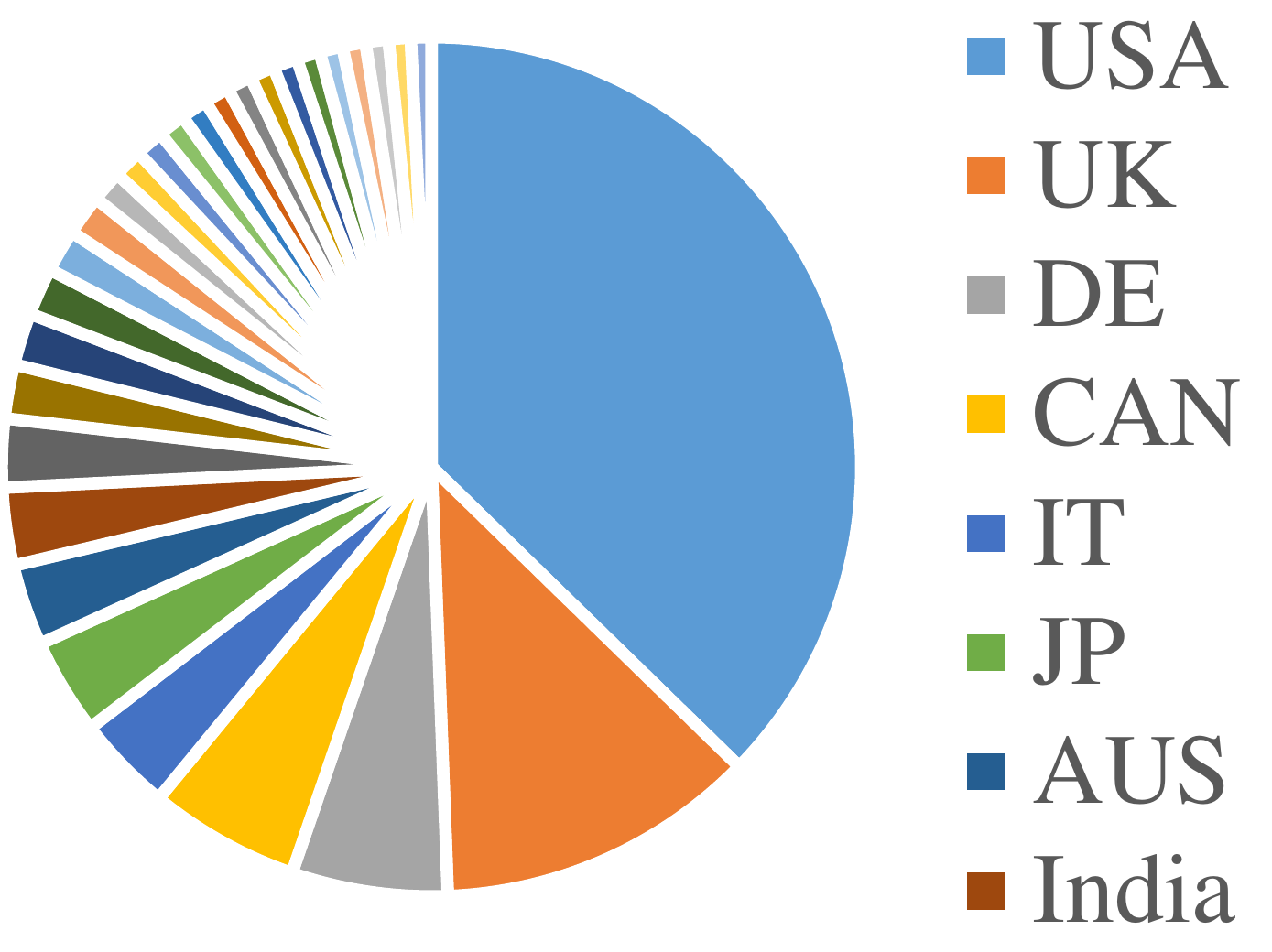}}
\subfigure[Profession]{
\label{fig:allprofession}
\includegraphics[width=0.157\textwidth]{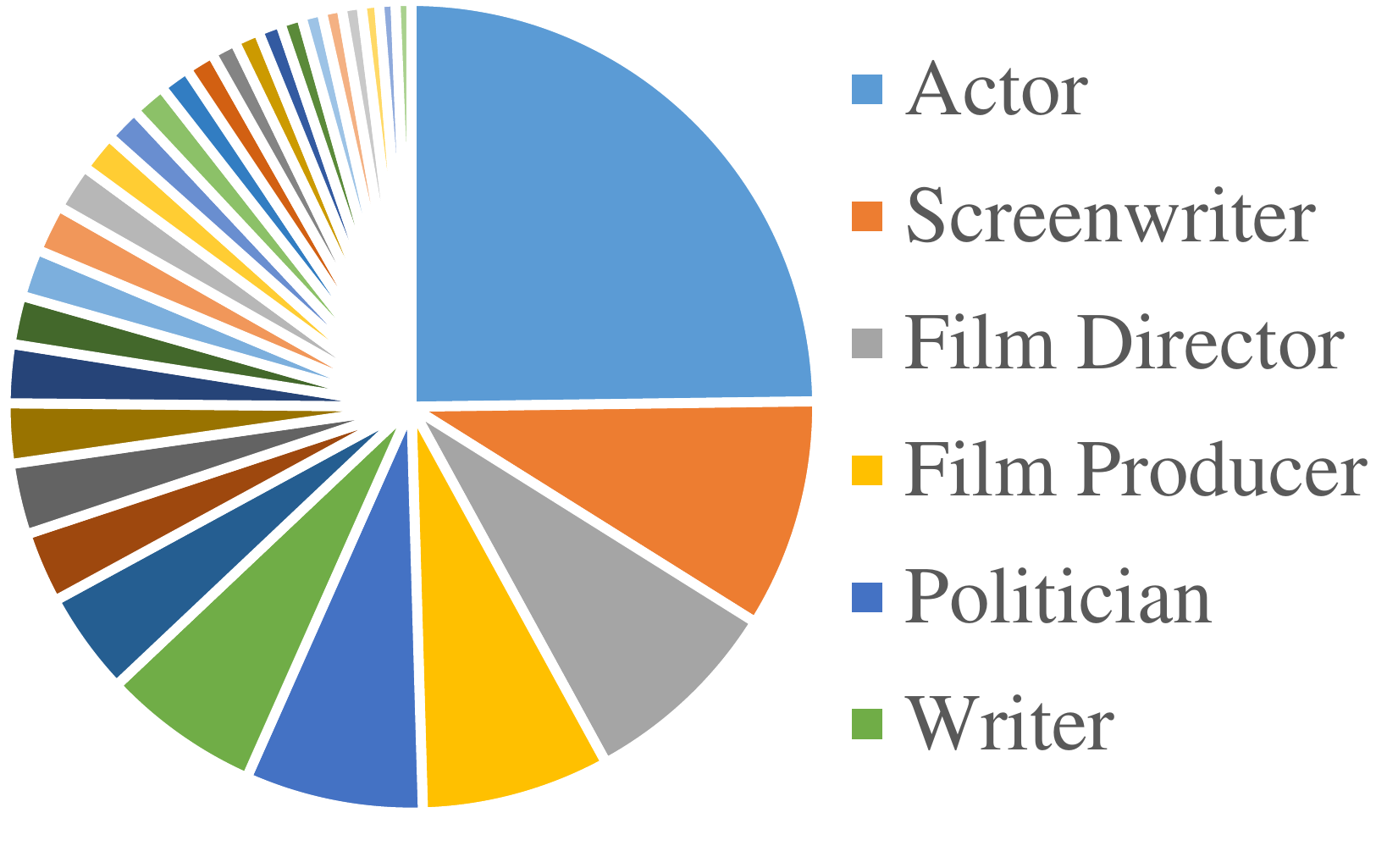}}
\vspace{-3mm}
\caption{Date of birth, nationality and profession of WebFace260M.}
\vspace{-4mm}
\label{fig:all_statistics}
\end{figure}

\noindent{\bf Face pre-processing.}
Faces are detected and aligned through five landmarks predicted by RetinaFace~\cite{RetinaFace}. For multi-face images, we only select the largest face with the above-threshold score, which can filter most improper faces (\eg background faces or wrong decoding). After pre-processing, there remains 4M identities/260M faces (WebFace260M) shown as Tab.\ref{table:training_set}. The statistics of WebFace260M are illustrated in Fig.\ref{fig:all_statistics} including date of birth, nationality and profession. Persons in WebFace260M come from more than 200 distinct countries/regions and more than 500 different professions with the date of birth back to 1846, which guarantees a great diversity in our training data.

\noindent{\bf Cleaned WebFace42M.}
We perform CAST pipeline (Sec.\ref{sec:cast}) to automatically clean the noisy WebFace260M and obtain a cleaned training set named WebFace42M, consisting of 42M faces of 2M subjects. Face number in each identity varies from 3 to more than 300, and the average face number is 21 per identity. As shown in Fig.\ref{fig:training_set} and Tab.\ref{table:training_set}, WebFace42M offers the largest cleaned training data for face recognition. Compared with the MegaFace2~\cite{MF2} dataset, the proposed WebFace42M includes 3 times more identities (2M vs. 672K), and near 10 times more images (42M vs. 4.7M). Compared with the widely used MS1M~\cite{MS1M}, our training set is 20 times (2M vs. 100K) and 4 times (42M vs. 10M) more in terms of  \# identities and \# photos. According to \cite{IMDB-Face}, there are more than 30\% and 50\% noises in MegaFace2 and MS1M, while noise ratio of WebFace42M is lower than 10\% (similar to CASIA-WebFace~\cite{CASIA-WebFace}) based on our sampling estimation. With such a large data size, we take a significant step towards closing the data gap between academia and industry.

\begin{figure}
\small
\centering
\subfigure[Pose]{
\label{fig:pose}
\includegraphics[width=0.13\textwidth]{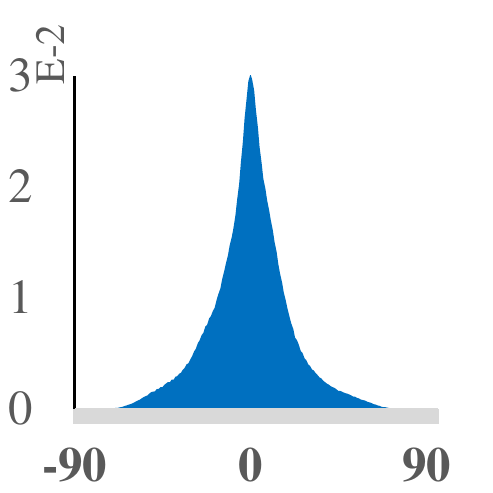}}
\subfigure[Age]{
\label{fig:age}
\includegraphics[width=0.13\textwidth]{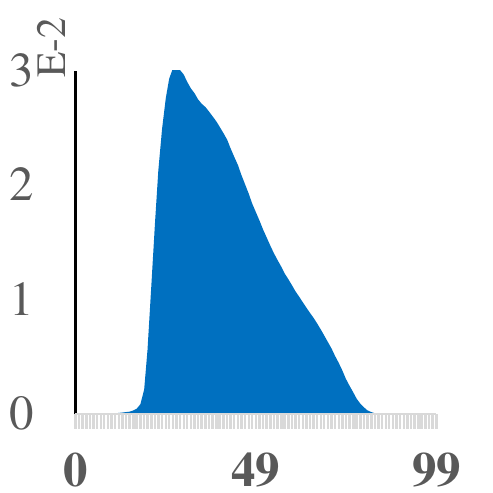}}
\subfigure[Race]{
\label{fig:race}
\includegraphics[width=0.19\textwidth]{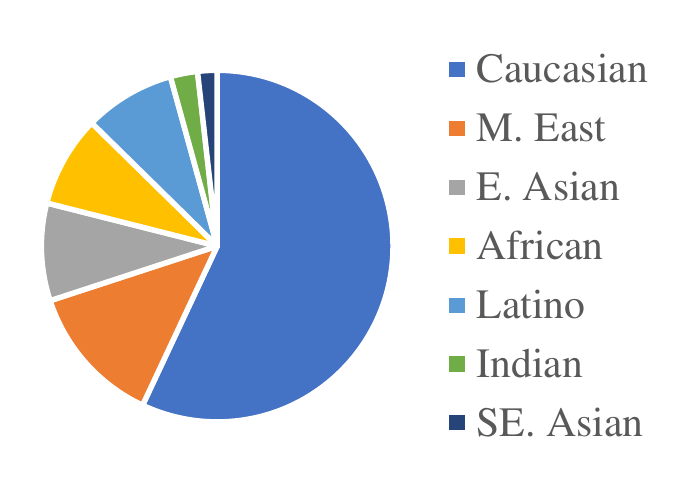}}
\vspace{-3mm}
\caption{Pose (yaw), age and race of WebFace42M.}
\vspace{-4mm}
\label{fig:face_attributes}
\end{figure}

\noindent{\bf Face attributes on WebFace42M.}
We further provide 7 face attribute annotations for WebFace42M, including pose, age, race, gender, hat, glass, and mask. Fig.\ref{fig:face_attributes} presents the distribution of our cleaned training data in different aspects. WebFace42M covers a large range of poses (Fig.\ref{fig:pose}), ages (Fig.\ref{fig:age}) and most major races in the world (Fig.\ref{fig:race}).

\section{Cleaning Automatically by Self-Training}
\label{sec:cast}

Since the images downloaded from the web are considerably noisy, it is necessary to perform a cleaning step to obtain high-quality training data. Original MS1M~\cite{MS1M} does not perform any dataset cleaning, resulting in near 50\% noise ratio, and significantly degrades the performance of the trained models. VGGFace~\cite{VGGFace}, VGGFace2~\cite{VGGFace2} and IMDB-Face~\cite{IMDB-Face} adopt semi-automatic or manual cleaning pipelines, which require expensive labor efforts. It becomes challenging to scale up the current annotation size to even more identities. Although the purification in MegaFace2~\cite{MF2} is automatic, its procedure is complicated and there are considerably more than 30 \% noises~\cite{IMDB-Face}. Another relevant exploration is to cluster faces via unsupervised approaches~\cite{otto2017clustering,kmeans,sibson1973slink} and supervised graph-based algorithm~\cite{CDP,GCND,GCNV,guo2020density,wang2019linkage}. However, these methods assume the whole dataset is clean, which is not suitable for the extremely noisy WebFace260M.

Recently, self-training~\cite{noisy_student, yalniz2019billion,parthasarathi2019lessons,radosavovic2018data}, a standard approach in semi-supervised learning~\cite{scudder1965probability,yarowsky1995unsupervised}, is explored to significantly boost the performance of image classification. Different from close-set ImageNet classification~\cite{ImageNet}, directly generating pseudo labels on open-set face recognition is impractical. Considering this inherent limitation, we carefully design the pipeline of Cleaning Automatically by Self-Training (CAST). Our first insight is performing self-training on open-set face recognition data, which is a scalable and efficient cleaning approach. Secondly, we find embedding feature matters in cleaning large-scale noisy face data.


\begin{figure}[t]
\centering
\includegraphics[width=1\linewidth]{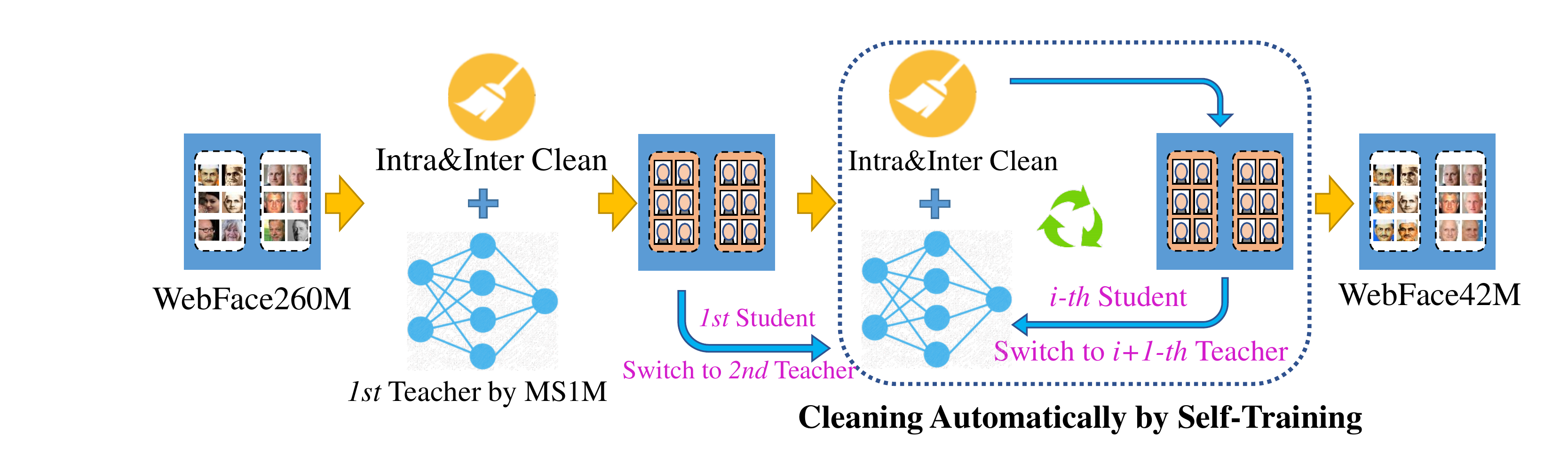}
\vspace{-2mm}
\caption{The proposed Cleaning Automatically by Self-Training (CAST). Firstly, an initial teacher trained with MS1MV2 is utilized to clean WebFace260M. Then a student model is trained on cleaned WebFace data. The CAST is performed by switching the student as the teacher until high-quality 42M faces are obtained. Every intra-class and inter-class cleaning is conducted on initial WebFace260M utilizing different teacher model.}
\label{fig:self_training}
\vspace{-4mm}
\end{figure}

The overall CAST framework is shown in Fig.\ref{fig:self_training}.
Following the self-training pipeline, (1) a teacher model (ResNet-100~\cite{ResNet}, ArcFace~\cite{ArcFace}) is trained with the public dataset (MS1MV2~\cite{ArcFace}) to clean the original 260M images, which mainly consists of intra-class and inter-class cleaning. (2) A student model (also ResNet-100, ArcFace) is trained on cleaned images from (1). Since the data size is much larger, this student generalizes better than the teacher. (3) We iterate this process by switching the student as the teacher until high-quality 42M faces are obtained. It is worth noting that each intra and inter class cleaning is conducted on initial WebFace260M by different teacher model.

\noindent{\bf Intra-class and inter-class cleaning.}
Since WebFace260M contains various noises such as outliers in a folder and identity overlaps between folders, it is impractical to perform unsupervised or supervised clustering on the whole dataset. Based on the observation that the image search results from Google are sorted by relevance and there is always a dominant subject in each search, the initial folder structure provides strong priors to guide the cleaning strategy: one folder contains a dominant subject and different folders may contain considerable overlapped identities.

Following these priors, we perform dataset cleaning by a two-step procedure: Firstly, face clustering is parallelly conducted in 4M folders (subjects) to select each dominant identity.  Specifically, for each face in a folder, 512-dimensional embedding feature is extracted by the teacher model, and then DBSCAN \cite{ester1996density} is utilized to cluster faces in this folder. Only largest cluster (more than 2 faces) in each fold is reserved. We also investigate other different designs of intra-class cleaning including GCN-D~\cite{GCND} and GCN-V~\cite{GCNV} in Sec.\ref{Comparisons_of_Data_Cleaning}. Secondly, we compute the feature center of each subject to perform inter-class cleaning. Two folders are merged if their cosine similarity is higher than 0.7, and the folder containing fewer faces would be deleted when the cosine similarity is between 0.5 and 0.7.

\begin{figure}
\small
\centering
\subfigure[Initial]{
\label{fig:0_score}
\includegraphics[width=0.11\textwidth]{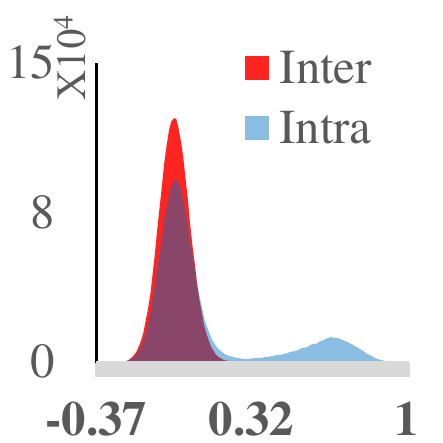}}
\subfigure[1st-iter]{
\label{fig:1_score}
\includegraphics[width=0.11\textwidth]{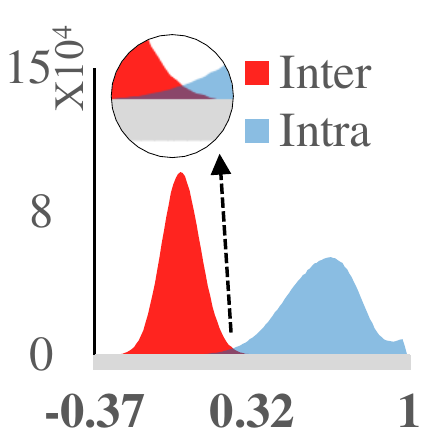}}
\subfigure[2nd-iter]{
\label{fig:2_score}
\includegraphics[width=0.11\textwidth]{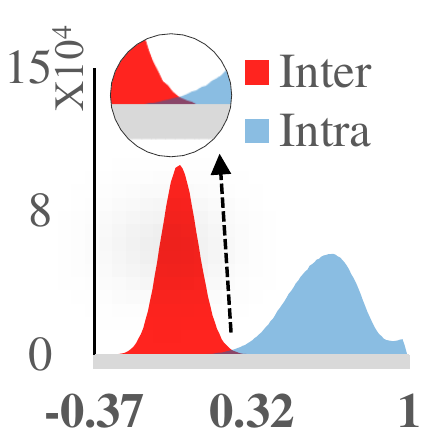}}
\subfigure[3rd-iter]{
\label{fig:3_score}
\includegraphics[width=0.11\textwidth]{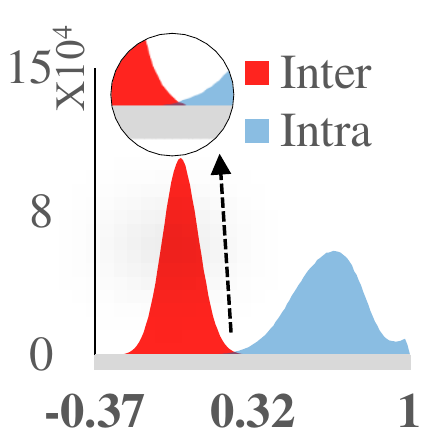}}
\vspace{-2mm}
\caption{Inter and intra class similarity distributions during different stages. Since initial folders are very noisy, score distributions are severely overlapped. Cleaner training set is obtained after more iterations. 100K folders are randomly selected here for showing the statistic changes during iterations.}
\vspace{-2mm}
\label{fig:intrainterclean}
\end{figure}

The effectiveness of the above intra-class and inter-class cleaning heavily depends on the quality of the embedding feature, which is guaranteed by the proposed self-training pipeline. The ArcFace model trained on MS1MV2 with ResNet-100 provides a good initial embedding feature to perform first round cleaning for WebFace260M. Then, this feature is significantly enhanced with more training data in later iterations. Fig.\ref{fig:intrainterclean} illustrates the score distribution during different stages of CAST, which indicates cleaner training set after more iterations. Furthermore, ablation study in Tab.\ref{tab:castcleaningresult} also validates the effectiveness of CAST pipeline. It is worth noting that the proposed CAST pipeline is compatible with any intra-class and inter-class strategies.

\noindent{\bf Remove duplicates and test set overlaps.}
After CAST, duplicates of each subject are removed when their cosine similarity is higher than 0.95. Furthermore, the feature center of each subject is compared with popular benchmarks (\eg LFW families~\cite{LFW,CALFW,CPLFW}, FaceScrub~\cite{FaceScrub}, IJB-C~\cite{IJB-C} etc.)
and the proposed test set in Sec.\ref{sec:test_set}, and overlaps are removed if the cosine similarity is higher than 0.7.

\begin{table}[t]
\begin{center}{\scalebox{0.8}{
\begin{tabular}{l|c|c|c}
\hline
\multicolumn{2}{c|}{Stages} & \# Identities & \# Faces \\ \hline
\hline
\multicolumn{2}{c|}{Collect name list and images} & 4,073,509 & 265,777,598   \\
\multicolumn{2}{c|}{Face pre-processing} & \textbf{4,008,130} & \textbf{260,890,076} \\ \hline
\multirow{2}{*}{First iteration}
& Intra-class & 3,341,761 & 61,792,387 \\
& Inter-class & 2,437,140 & 50,672,354 \\ \hline
\multirow{2}{*}{Second iteration}
& Intra-class & 3,027,814 & 60,274,892 \\
& Inter-class & 2,176,427 &47,352,741 \\ \hline
\multirow{2}{*}{Third iteration}
& Intra-class & 2,878,886 & 58,155,345 \\
& Inter-class &2,070,870  & 46,220,417 \\ \hline
\multicolumn{2}{c|}{Remove duplicates} & 2,070,870 &43,977,802 \\
\multicolumn{2}{c|}{Remove test set overlaps} & \textbf{2,059,906} & \textbf{42,474,558}\\ \hline
\end{tabular}}}
\end{center}
\vspace{-4mm}
\caption{ The \# identities and \# images statistics during different stages.}
\label{table:cast_sta}
\vspace{-4mm}
\end{table}

\noindent{\bf Dataset statistics.}
The statistics of \# identities and \# images during different stages are shown in Tab.\ref{table:cast_sta}. After face pre-processing for downloaded images, there are 4,008,130 identities and 260,890,076 faces (WebFace260M). The face set becomes cleaner under more CAST iterations, which results in fewer identities and faces. Finally, we obtain 2,059,906 identities and 42,474,558 faces (WebFace42M) after removing duplicates and test set overlaps.

\section{FRUITS Protocol}

\subsection{Evaluation Protocol}

Popular evaluation protocols for face recognition mainly target the pursuit of accuracy. For example, CFP~\cite{CFP}, AgeDB~\cite{AgeDB}, CALFW~\cite{CALFW} and CPLFW~\cite{CPLFW} evaluate the verification accuracy under different intra-class variations (\eg pose and age). MegaFace~\cite{MegaFace} and IJB-C~\cite{IJB-C} serve for both accuracy of large-scale face verification and identification. YTF~\cite{YTF} and IQIYI-Video~\cite{IQIYI2018} compare the accuracy of video-based verification. Different model ensemble and post-processing \cite{PFE} could be adopted for higher performance under these protocols. However, face recognition in real-world application scenarios is always restricted by inference time.

Recently, lightweight face recognition challenge \cite{LFR} takes a step toward this goal by constraining the FLOPs and model size of submissions. Since different neural network architectures can be quite different in terms of real inference times, this protocol is not a straightforward solution. Furthermore, it does not consider face detection and alignment, which are prerequisite components in most modern face recognition systems. To the best of our knowledge, NIST-FRVT~\cite{FRVT} is the only benchmark employing the time-constrained protocol. However, strict submission policy (participants can only send one submission every four calendar months) hinders researchers to freely evaluate their algorithms.

In this paper, we design the Face Recognition Under Inference Time conStraint (FRUITS) protocol, which enables academia to comprehensively evaluate their face matchers. Referring to~\cite{FRVT},  inference time is measured on a single core of an Intel Xeon CPU E5-2630-v4@2.20GHz processor. Considering different application scenarios, FRUITS protocol sets a series of tracks:

\noindent{\bf {FRUITS-100}}: The whole face recognition system must distinguish image pairs within 100 milliseconds, including pre-processing (\eg face detection and alignment), feature embedding for recognition, and matching. FRUITS-100 track targets on evaluating lightweight face recognition system which can be deployed on mobile devices.

\noindent{\bf{FRUITS-500}}: This track follows FRUITS-100 setting, except that time constraint is increased to 500 milliseconds. This track aims to evaluate modern and popular networks deployed in the local surveillance system.

\noindent{\bf{FRUITS-1000}}: Following NIST-FRVT, FRUITS-1000 adopts time constraint of 1000 milliseconds and aims to compare capable recognition models performed on clouds.

\subsection{Test Set}
\label{sec:test_set}

\begin{table}[t]
\begin{center}{\scalebox{0.75}{
\begin{tabular}{c|c|c|c|c|c}
\hline
\multicolumn{2}{c|}{Attributes}& \# Id. & \# Faces & \# Impostor & \# Genuine \\ \hline
\hline

\multicolumn{2}{c|}{\textbf{All}} & \textbf{2,225} & \textbf{38,578} & \textbf{743,683,994} & \textbf{427,759}\\\hline

\multirow{2}{*}{\tabincell{l}{Age}}

&Cross-age-10& - & - & 374,849,719 & 109,350\\
&Cross-age-20& - & - & 196,770,680 & 27,056 \\ \hline

\multirow{4}{*}{\tabincell{l}{Race}}

&Caucasian& 997 & 17,462  &76,747,746  &138,454  \\
&East Asian&647  & 12,401   & 20,384,596 & 60,219 \\
&African& 441 & 6,395 & 2,666,162 & 23,878 \\
&Others&140 & 2,320 & - & - \\ \hline

\multirow{2}{*}{\tabincell{l}{Gender}}

&Male& 1,370 & 22,846 & 260,724,139 & 234,296  \\
&Female& 855 & 15,732 & 123,546,583 & 193,463 \\ \hline

\multirow{3}{*}{\tabincell{l}{Scenarios}}

& Controlled& - & 20,446 & 208,876,619  &132,616 \\
&Wild& - &  18,132   & 164,250,414 & 125,232\\
&Cross-scene& - & - &370,556,961  & 169,911   \\  \hline

\end{tabular}}}
\end{center}
\vspace{-4mm}
\caption{The statistics of our test set. - means corresponding statistics or comparisons are omitted.}
\vspace{-4mm}
\label{table:test_set_num}
\end{table}

Since public evaluations are most saturated and may contain noise,
we manually construct an elaborated test set for FRUITS. It is well known that recognizing strangers, especially when they are similar-looking, is a difficult task even for experienced vision researchers. Therefore, our multi-ethnic annotators only select their familiar celebrities, which ensure the high-quality of the test set.
Besides, annotators are encouraged to gather attribute-balanced faces, and recognition models are introduced to guide hard sample collection.
 The statistics of the final test set are listed in Tab.\ref{table:test_set_num}. In total, there are 38,578 faces of 2,225 identities. Rich attributes (\eg age, race, gender, controlled or wild) are accurately annotated. In the future, we will actively maintain and update this test set.

\subsection{Metrics}

Based on the proposed FRUITS protocol and test set, we perform 1:1 face verification across various attributes. Tab.\ref{table:test_set_num} shows numbers of imposter and genuine in different verification settings. \emph{All} means impostors are paired without attention to any attribute, while later comparisons are conducted on age, race, gender and scenario subsets. \emph{Cross-age} refers to cross-age (more than 10 and 20 years) verification, while \emph{Cross-scene} means pairs are compared between controlled and wild settings. Different algorithms are measured on False Non-Match Rate (FNMR) \cite{FRVT}, which is defined as the proportion of mated comparisons below a threshold set to achieve the False Match Rate (FMR) specified. FMR is the proportion of impostor comparisons at or above that threshold. \textbf{Lower FNMR at the same FMR is better}.

\section{Experiments of Million-level Recognition}

\subsection{Implementation Details}

In order to fairly evaluate the performance of different face recognition models, we reproduce representative algorithms (\ie CosFace \cite{CosFace}, ArcFace \cite{ArcFace} and CurricularFace \cite{Curricularface}) in one Gluon codebase with the hyper-parameters referred to the original papers. Default batch size per GPU is set as 64 unless otherwise indicated. Learning rate is set as 0.05 for a single node (8 GPUs), and follows the linear scaling rule \cite{goyal2017accurate} for the training on multiple nodes (\ie $0.05\times$\# machines). We decrease the learning rate by 0.1$\times$ at 8, 12, and 16 epochs, and stop at 20 epochs for all models. During training, we only adopt the flip data augmentation.

\subsection{Distributed Training}

When using the large-scale WebFace42M as the training data and computationally demanding backbones as the embedding networks, the model training can take several weeks on one machine. Such a long training time makes it difficult to efficiently perform experiments. Inspired by the distributed optimization on ImageNet~\cite{goyal2017accurate}, we apportion the workload of model training to clusters. To this end, parallel on both feature $X$ and center $W$,  mixed-precision (FP16) and large-batch training are adopted in this paper.

\begin{figure}[t]
\centering
\vspace{-0.2cm}
\includegraphics[width=0.90\linewidth]{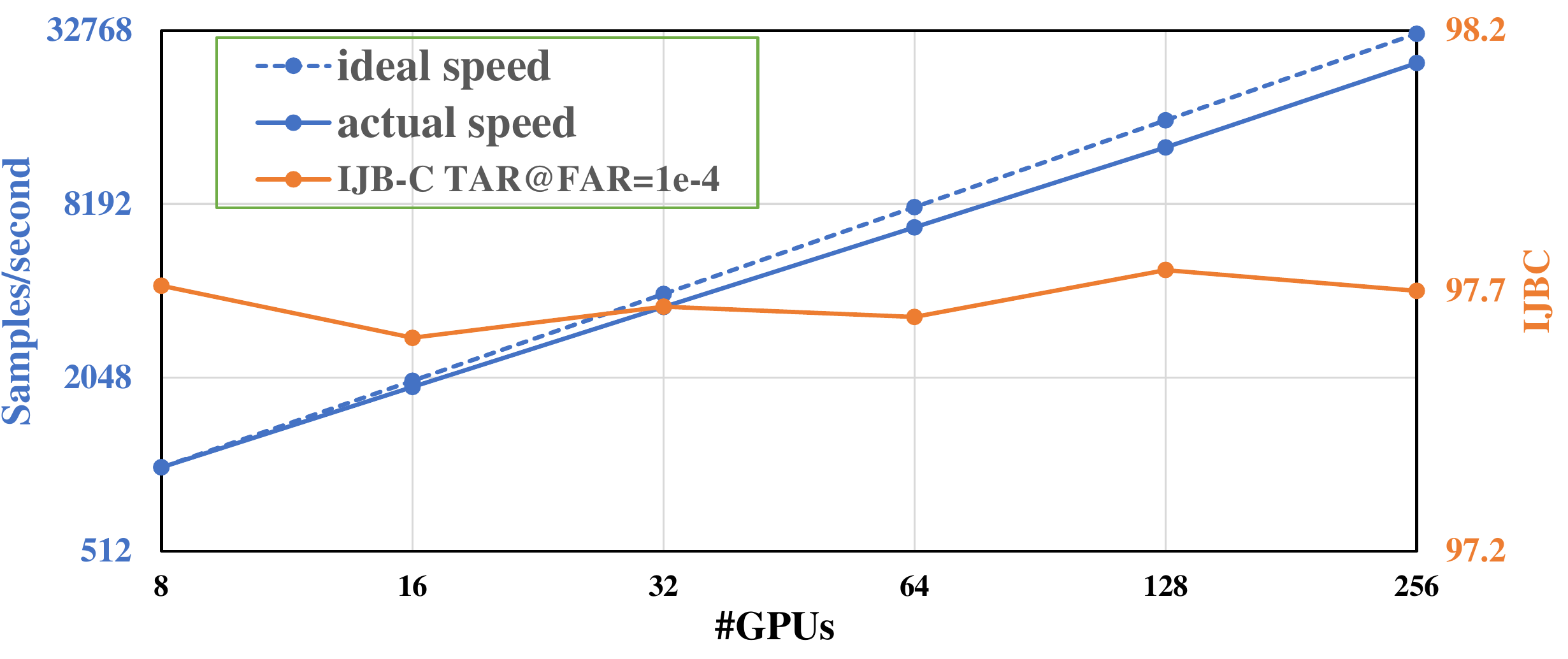}
\caption{Speed and performance of our distributed training system. The proposed system can almost linearly accelerate the training with comparable performance. 100\% data (WebFace42M) is used in these experiments.}
\label{fig:distributed_training}
\vspace{-0.1cm}
\end{figure}

Speed and performance of our distributed training system are illustrated in Tab.\ref{table:distributed_training} and Fig.\ref{fig:distributed_training}. Parallelization on both feature $X$ and center $W$ as well as mixed-precision (FP16) significantly reduce the consumption of GPU memory and speed up the training process, while similar performance can be achieved. Equipped with 8 nodes (64 GPUs), the training speed is scaled to 12K samples/s and 11K samples/s on WebFace4M (10\% data) and WebFace12M (30\% data), respectively. The corresponding training time is only 2 hours and 6 hours. Furthermore, the scaling efficiency of our training system is above 80\% when applied to large-scale WebFace42M on 32 nodes (256 GPUs). Therefore, we can reduce the training time of the ResNet-100 model from 233 hours (1 node) to 9 hours (32 nodes) with comparable performance.

\begin{table}[ht]
\begin{center}{\scalebox{0.72}{
\begin{tabular}{l|c|c|c|c|c|c}
\hline
 Data & B$\times$G$\times$M &FP32/16 &Parallel & Speed  & Time & IJB-C \\
\hline
\hline
\multirow{6}{*}{\tabincell{c}{10\%}}
& 32$\times$8$\times$1 & FP32 & $X$ (7913)&0.6K& 39h & 96.67 \\
& 64$\times$8$\times$1 & FP32 & $X$ $W$ (7521)  &0.9K&26h& 96.83 \\
& 64$\times$8$\times$1 & FP16 & $X$ (7551) &1K& 23h & 96.80 \\
& 64$\times$8$\times$1 & FP16 & $X$ $W$ (7182)&1.8K& 13h & 96.78 \\
& 64$\times$8$\times$4 & FP16 & $X$ $W$ (7125) &6.3K& 4h  & 96.73 \\
& 64$\times$8$\times$8 & FP16 & $X$ $W$ (7119) &12.4K& 2h  & 96.77 \\
\hline
\multirow{3}{*}{\tabincell{c}{30\%}}
& 64$\times$8$\times$1 & FP16 & $X$ $W$ (8901)&1.7K&41h & 97.41\\
& 64$\times$8$\times$4 & FP16 & $X$ $W$ (8519)&5.5K&13h &97.50\\
& 64$\times$8$\times$8 & FP16 & $X$ $W$ (8455)&11.3K&6h &97.47 \\
\hline
\multirow{4}{*}{\tabincell{c}{100\%}}
& 32$\times$8$\times$1 & FP16 & $X$ $W$ (10503)&1K& 233h & 97.71 \\
& 32$\times$8$\times$8 & FP16 & $X$ $W$ (8359)&6.8K&34h & 97.65\\
& 32$\times$8$\times$16 & FP16 & $X$ $W$ (8297)&12.9K&18h & 97.74\\
& 32$\times$8$\times$32 & FP16 & $X$ $W$ (8221)&25.3K& \textbf{9h}& 97.70 \\
\hline
\end{tabular}}}
\end{center}
\vspace{-0.5cm}
\caption{Speed and performance comparison of distributed training. ArcFace using ResNet-100 is adopted. B, G and M refer to batch size per GPU, \# GPUs per machine, and \# machines. $X$ and $W$ mean feature and center, and numbers in bracket are the GPU memory usage (MB). Performance is reported on IJB-C (TAR@FAR=1e-4).}
\label{table:distributed_training}
\vspace{-0.5cm}
\end{table}

\subsection{Comparisons of Training Data}

For comprehensively analysing the influence of training data, the proposed WebFace42M is compared with public counterparts including MS1M families \cite{MS1M,deng2017marginal,ArcFace,glintweb},  MegaFace2 \cite{MF2} and IMDB-Face \cite{IMDB-Face}. 10\% (WebFace4M) and 30\% (WebFace12M) random selection of our full data are also employed for further analysing of the training data. The statistics of different training sets are illustrated in Tab.\ref{table:training_set}. Evaluation sets used in this experiment include popular verification sets (\eg LFW \cite{LFW}, CFP-FP \cite{CFP}, CPLFW \cite{CPLFW}, AgeDB \cite{AgeDB} and CALFW \cite{CALFW}), RFW \cite{RFW}, MegaFace \cite{MF2}, IJB-C \cite{IJB-C} and our test set.

\begin{figure}[t]
\centering
\includegraphics[width=0.95\linewidth]{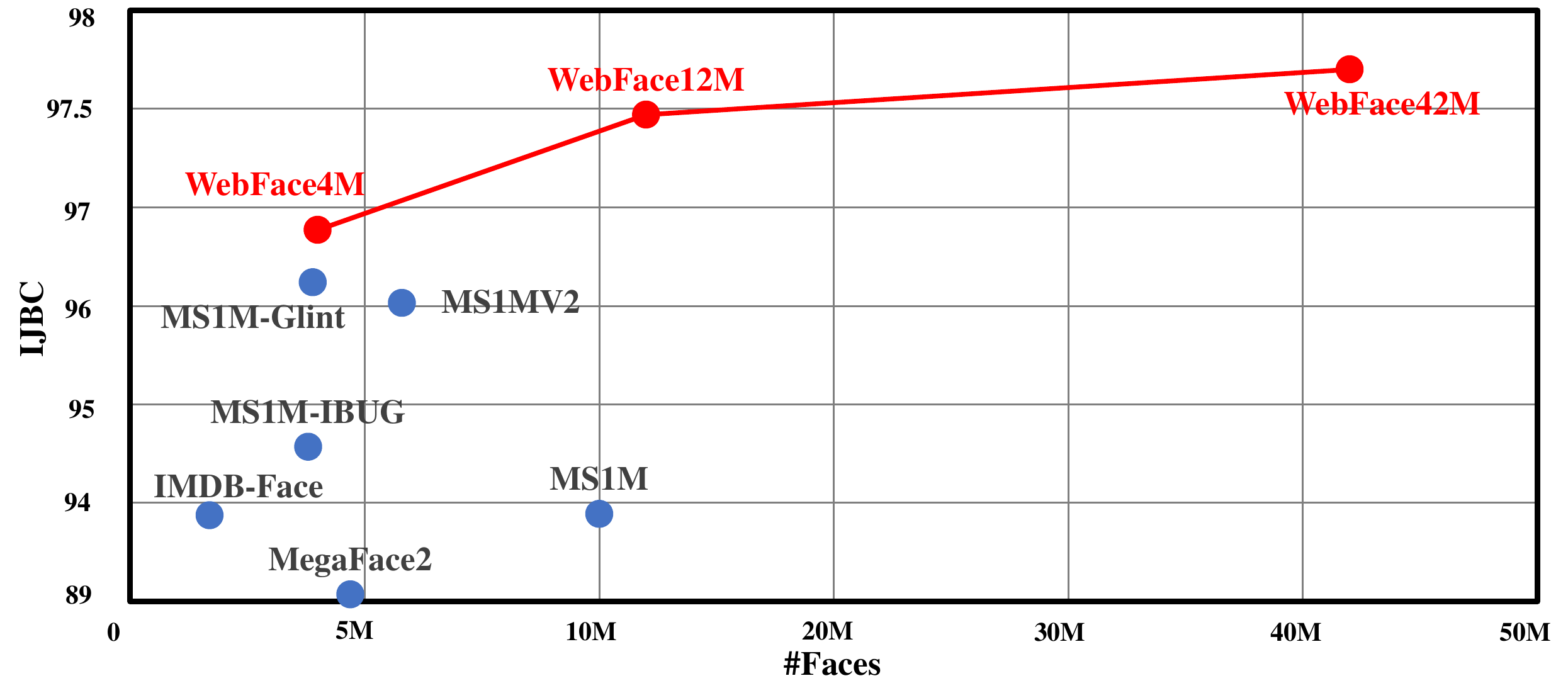}
\caption{Performance of ArcFace models (ResNet-100) trained on the WebFace envelopes counterparts trained on the public training data.}
\vspace{-3mm}
\label{fig:training_setvs}
\end{figure}

As we can see from Tab.\ref{tab:trainingdatavs} and Fig.\ref{fig:training_setvs}, the proposed WebFace42M breaks the bottleneck of training data for deep face recognition across various loss functions and test sets. Specifically, WebFace42M reduces relative 40\% error rate on the challenging IJB-C dataset compared with MS1MV2, boosting TAR from 96.03\% to 97.70\% @10-4 FAR. Along with the increment of data scale (\ie 10\%, 30\%, and 100\%), there exists a consistent improvement in performance as observed in Fig.\ref{fig:training_setvs}. On our test set, the relative promotion is near 70\% when trained on WebFace42M.
Impressively, the models trained on 10\% data, WebFace4M, achieve superior performance compared to models trained on  MS1M families and MegaFace2, which include even more \# faces. Undisputedly, the training data comparison confirms the effectiveness and necessity of our WebFace42M in levelling playing field for million-scale face recognition.

Besides reporting the results of ResNet-100, we also train ArcFace models by using a smaller network, ResNet-14, on different portions of our data (\ie 10\%, 30\% and 100\%). As given in Tab.\ref{table:small_model}, there is also a consistent performance gain for ResNet-14 when more training data are progressively employed. Therefore, the proposed WebFace42M is not only beneficial to the large model (\eg ResNet-100) but also valuable for the lightweight model.

\begin{table}[t]
\centering
\begin{center}{\scalebox{0.72}{
\begin{tabular}{c|c|c|c|c|c|c}
\hline
Data & Loss & Pairs &RFW & Mega &IJB-C & Our test $\downarrow$ \\
\hline \hline

\multirow{3}{*}{MS1M}
& CosFace    &95.69 &98.09& 96.21 &92.96 & 26.87   \\
& ArcFace   &95.53 &97.64& 97.67 & 93.45 &  19.47 \\
& Curricular   & 95.71  &98.12 & 96.86 & 92.99 &  33.14 \\
\hline

\multirow{3}{*}{MS1M-IBUG}
& CosFace    &95.67 &97.62& 97.33 & 94.35 &  6.36 \\
& ArcFace    &95.49 &97.78& 97.27   &  94.57  & 7.05\\
& Curricular   & 95.71& 97.86& 97.19 & 94.72 & 7.13 \\
\hline
\multirow{3}{*}{MS1MV2}
& CosFace    &97.05 &98.85& 98.30& 96.01 &  4.49 \\
& ArcFace   &97.10 &98.98&98.40&   96.03 & 5.08\\
& Curricular   & 97.23& 99.02&98.46 &  96.21 & 4.95\\
\hline
\multirow{3}{*}{MS1M-Glint}
& CosFace    &95.99 &99.59&98.60& 96.15 &  6.11 \\
& ArcFace    &95.81 &99.60&98.48& 96.24 &  6.66\\
& Curricular   &96.41 & 99.65&98.57& 96.31  & 6.93 \\
\hline

\multirow{3}{*}{MegaFace2}
& CosFace   &92.52 &88.90&86.62& 87.75 & 45.90   \\
& ArcFace   &93.18 &89.45&88.28&  89.35  & 41.58\\
& Curricular   & 93.40&90.06 & 88.32 & 90.11  & 41.97\\
\hline

\multirow{3}{*}{IMDB-Face}
& CosFace     &96.41 &93.80&94.03& 93.96  &16.73 \\
& ArcFace   &96.40 &93.08&93.48&  93.37 & 19.07 \\
& Curricular   &96.62 &94.11&93.63 &  94.12 & 19.23\\
\hline

\multirow{3}{*}{WebFace4M}
& CosFace   &97.37 &98.16&97.59& 96.86  &  4.43 \\
& ArcFace   &97.39 &98.14&97.60&96.77  &  4.95 \\
& Curricular   & 97.40 &98.14&97.94& 97.02  & 4.33 \\
\hline

\multirow{3}{*}{WebFace12M}
& CosFace   & 97.61&99.15&98.66 &97.41  & 2.16 \\
& ArcFace   &97.66 &99.08&98.82 & 97.47 & 2.34 \\
& Curricular   &97.68 &99.18&98.75 &97.51  & 2.44 \\

\hline

\multirow{3}{*}{WebFace42M}

& CosFace  & 97.76 & 99.41&99.02& 97.68  & 1.72  \\
& ArcFace  &97.65 &99.33&99.02& 97.70 & 1.58    \\
& Curricular   & 97.68& 99.39& 99.11 &  97.76 &1.63 \\

\hline
\end{tabular}}}
\end{center}
\vspace{-4mm}
\caption{Performance (\%) of different training data. ResNet-100 backbone \textbf{without flip test} is adopted. Pairs refers to average accuracy on~\cite{LFW,CFP,AgeDB,CALFW,CPLFW}, RFW refers to average accuracy on~\cite{RFW}, Mega refers to rank-1 identification on \cite{MegaFace}, IJB-C is TAR@FAR=1e-4 on~\cite{IJB-C}. Last column is FNMR@FMR=1e-5 on \emph{All} pairs comparison of our test set.}
\label{tab:trainingdatavs}
\end{table}

\begin{table}[t]
\centering

\begin{center}{\scalebox{0.7}{
\begin{tabular}{c|c|c|c}
\hline
 Training data &WebFace4M & WebFace12M& WebFace42M  \\ \hline\hline
 IJB-C &93.13 &93.92 & 94.22    \\
\hline
\end{tabular}}}
\end{center}
\vspace{-4mm}
\caption{Performance of ArcFace models trained with ResNet-14 on different portions of WebFace42M. TAR@FAR=1e-4 on IJB-C is reported.}
\vspace{-4mm}
\label{table:small_model}
\end{table}

\subsection{Comparisons of Data Cleaning}
\label{Comparisons_of_Data_Cleaning}

As shown in Tab.\ref{tab:castcleaningresult}, the CAST pipeline is compared with other cleaning strategies on the original MS1M \cite{MS1M} and WebFace260M. Specifically, for MS1M results, the initial teacher model is trained on IMDB-Face \cite{IMDB-Face} by using ResNet-100 and ArcFace. Then, CAST is conducted on the noisy MS1M following Sec.\ref{sec:cast}. After steps of iteration, our fully automatic cleaning strategy provides purified data for model training, outperforming semi-automatic methods used in ~\cite{deng2017marginal,ArcFace,glintweb}. Compared with the most recent GCN-based cleaning \cite{MillionCelebs}, the data cleaned by the CAST also achieves higher performance.


\begin{table}[t]
\centering
\begin{center}{\scalebox{0.75}{
\begin{tabular}{c|c|c|c|c|c}
\hline
Data & \# Id & \# Face & Pairs & MegaFace & IJB-C \\\hline\hline
 MS1M &100K &10M & 95.53   &97.67 & 93.45\\\hline
 MS1M-IBUG& 85K &3.8M & 95.49   &97.27 & 94.57\\
 MS1MV2& 85K &5.8M & 97.10   &98.40 &96.03\\
 MS1M-Glint& 87K &3.9M & 95.81   & 98.48&96.24 \\
 MS1M-GCN \cite{MillionCelebs} & - & - & 96.51 & - & -  \\ \hline
 MS1M by CAST-1&94K &6.3M & 95.37 & 97.93  &94.31\\
 MS1M by CAST-2&92K &5.5M & 97.08 & 98.47  &95.90\\
 MS1M by CAST-3& 91K& 4.9M& 97.42 & 98.61  &96.55\\
 MS1M by CAST-4& 91K& 4.9M& 97.49 & 98.57 &96.52\\
\hline
 WebFace by CAST-1&2.4M &46M & 97.42 & 98.64  &97.28\\
 WebFace by CAST-2&2.1M &43M & 97.53 &  98.98 &97.51\\
 WebFace by CAST-3&2M   & 42M   & 97.65 & 99.02  &97.70\\
 WebFace by CAST-4&2M& 42M  & 97.69 & 99.08  &97.66\\
\hline
\end{tabular}}}
\end{center}
\vspace{-4mm}
\caption{Comparisons of CAST and other data cleaning pipelines. ResNet-100 using the ArcFace loss is adopted here. For our WebFace, different iterations are compared. CAST-1 means the first-round iteration.}
\label{tab:castcleaningresult}
\end{table}

\noindent{\bf Iterations of CAST.} Tab.\ref{tab:castcleaningresult} also shows the increasing data purity after more iterations in MS1M and WebFace260M. The accuracy gradually increases from 1st to 3rd iteration, while 4th iteration shows saturated performance. Therefore, we set the iteration number as 3 for CAST.

\noindent{\bf Intra-class Cleaning.} In this experiment, we compare different intra-class cleaning methods under the framework of CAST. Both unsupervised methods (\eg K-means \cite{kmeans} and DBSCAN \cite{ester1996density}) and supervised methods (\eg GCN-D \cite{GCND} and GCN-V \cite{GCNV}) are explored to find the dominant subject in each noisy folder. As shown in Tab.\ref{table:intra_cleaning}, DBSCAN achieves 96.55\% TAR@FAR=1e-4 on IJB-C, significantly outperforming K-Means (96.03\%) and slightly surpassing the supervised GCN-based strategies (96.48\%  for GCN-D and 96.42\% for GCN-V). As the GCN-based strategies can be sub-optimal for the extremely noisy folders, we finally select DBSCAN \cite{ester1996density} as our intra-class cleaning method.

\begin{table}[t]
\centering
\begin{center}{\scalebox{0.75}{
\begin{tabular}{c|c|c|c|c|c}
\hline
Data & \# Id & \# Face & Pairs & MegaFace & IJB-C \\\hline\hline


 K-Means &93K &5.2M & 95.17   &97.31 & 96.03\\
 DBSCAN& 91K &4.9M & {\bf 97.42}   & {\bf 98.61}&{\bf 96.55}\\
 \hline
 GCN-D& 86K &4.4M & 96.56   & 98.55& 96.48\\
 GCN-V& 82K &4.5M & 96.93   & 98.29&96.42 \\
\hline
\end{tabular}}}
\end{center}
\vspace{-2mm}
\caption{Comparisons of different intra-class cleaning methods for MS1M. ResNet-100 using the ArcFace loss is adopted here.}
\vspace{-4mm}
\label{table:intra_cleaning}
\end{table}

\subsection{Baselines under FRUITS Protocols}

In this section, we set up a series of baselines under the proposed FRUITS protocols. In Tab.\ref{table:time}, we illustrate different face recognition systems (including different module settings of face detection, alignment, feature embedding) and their inference time. In our baselines, representative network architectures are explored, covering MobileNet \cite{MobileNet,chen2018mobilefacenets}, EfficientNet \cite{tan2019efficientnet}, AttentionNet \cite{AttentionNet}, ResNet \cite{ResNet}, SENet \cite{SENet}, ResNeXt \cite{ResNeXt} and RegNet \cite{RegNet} families. All the models are trained on WebFace42M with ArcFace.

\begin{figure}
\small
\centering

\subfigure[{\scriptsize FMR-FNMR for FRUITS-100}]{
\label{fig:allfmr100}
\includegraphics[width=0.51\linewidth]{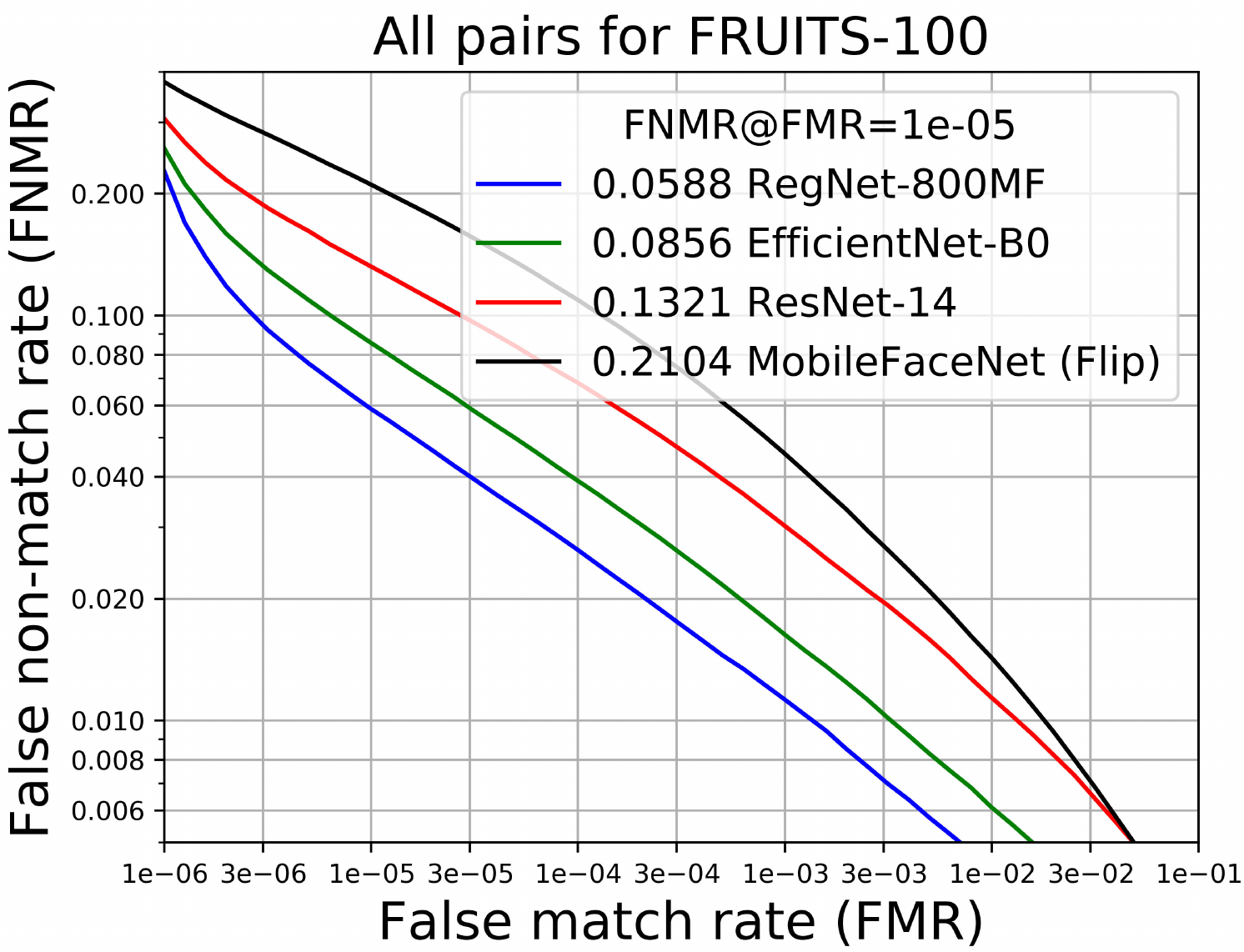}}
\subfigure[{\scriptsize Attributes for FRUITS-100}]{
\label{fig:allattribute100}
\includegraphics[width=0.43\linewidth]{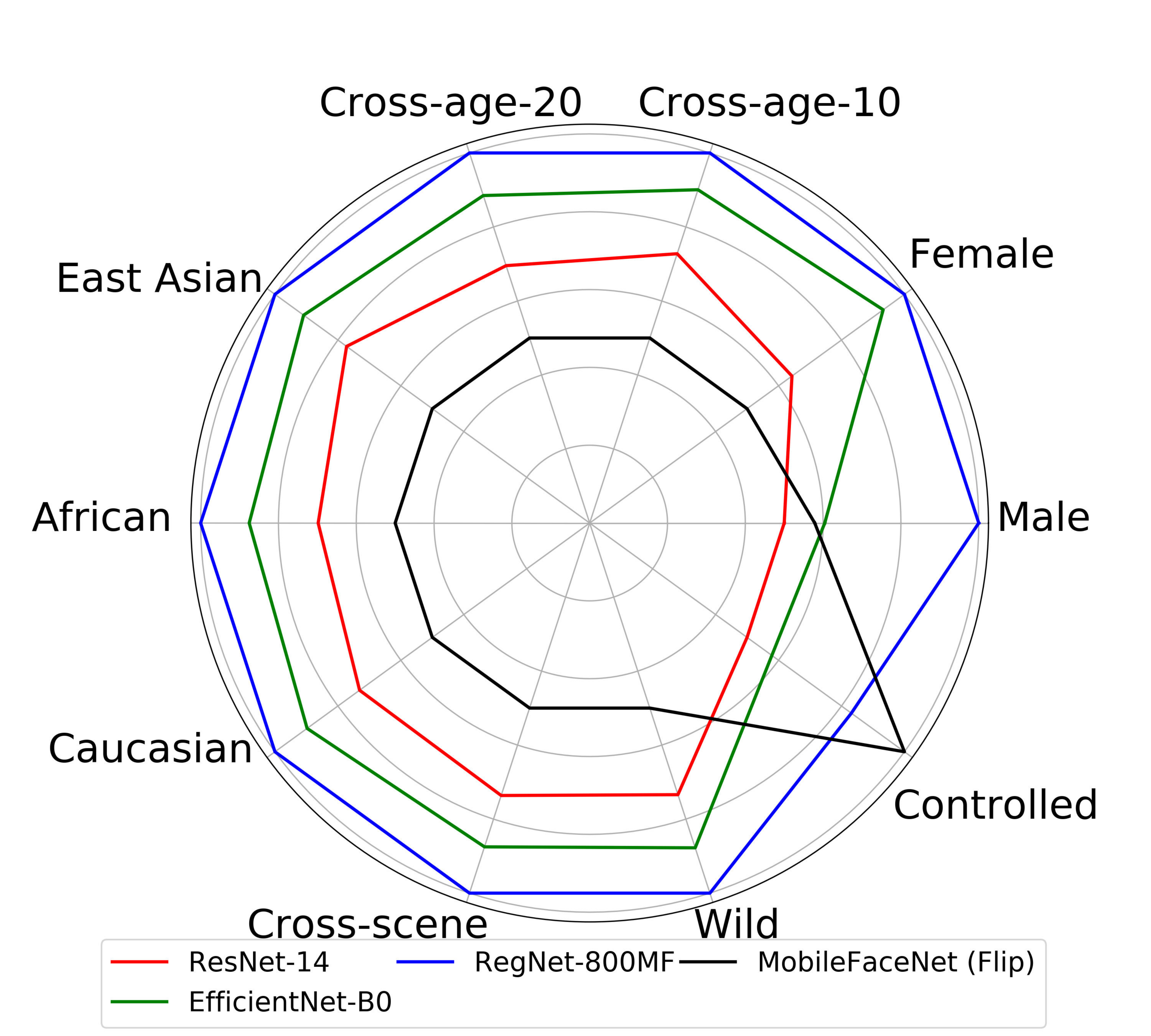}}

\subfigure[{\scriptsize FMR-FNMR for FRUITS-500}]{
\label{fig:allfmr500}
\includegraphics[width=0.51\linewidth]{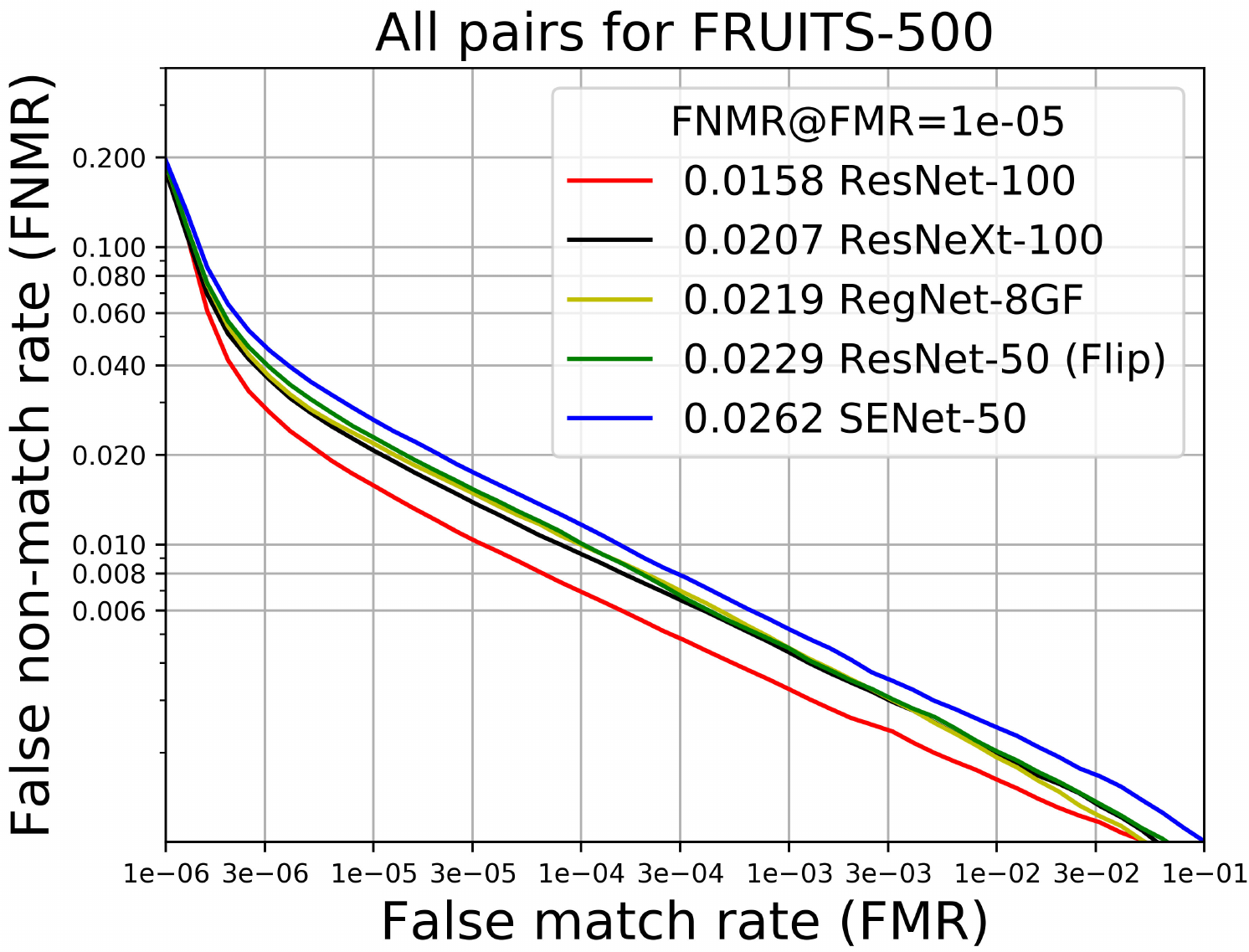}}
\subfigure[{\scriptsize Attributes for FRUITS-500}]{
\label{fig:allattribute500}
\includegraphics[width=0.43\linewidth]{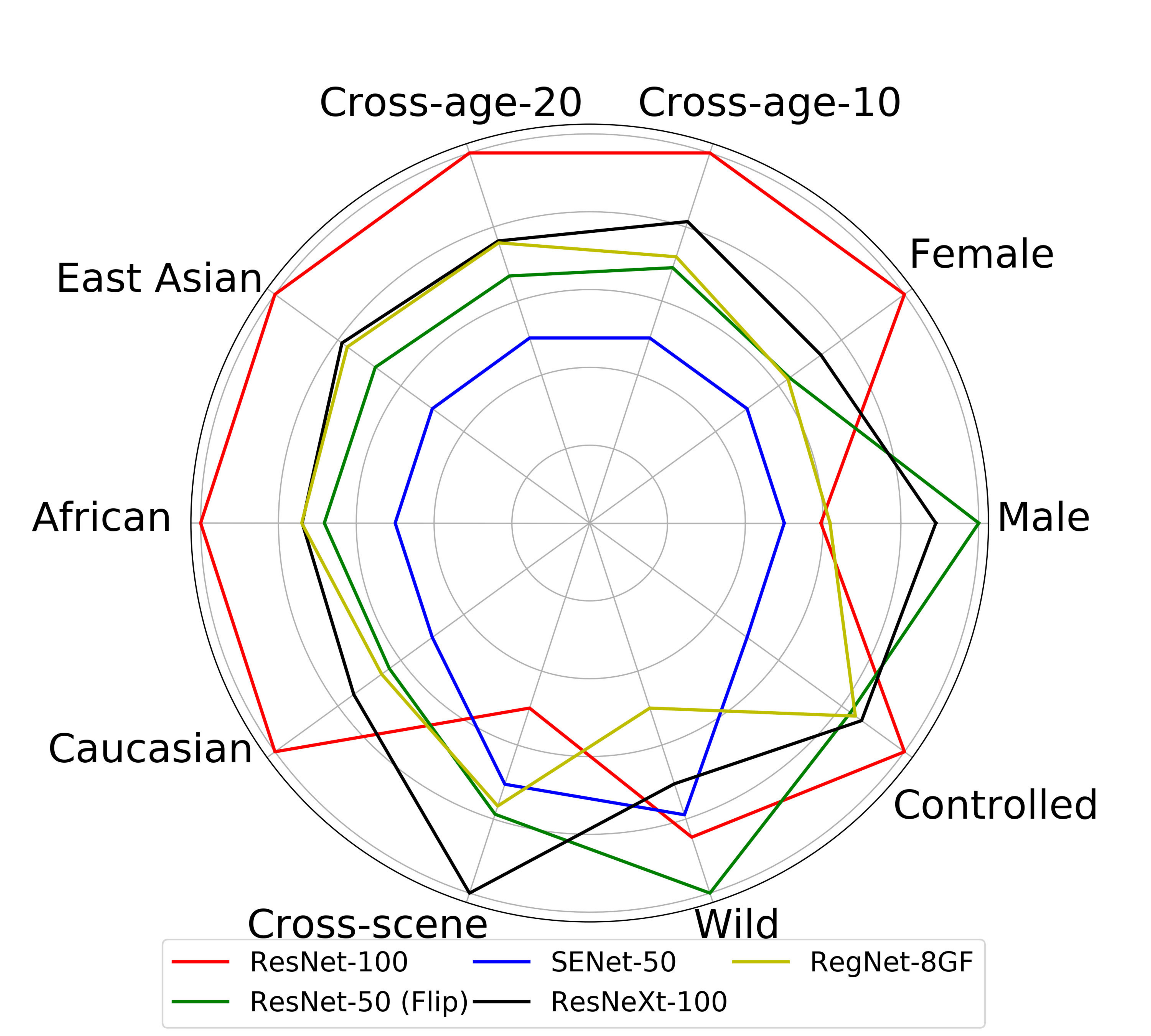}}

\subfigure[{\scriptsize FMR-FNMR for FRUITS-1000}]{
\label{fig:allfmr1000}
\includegraphics[width=0.51\linewidth]{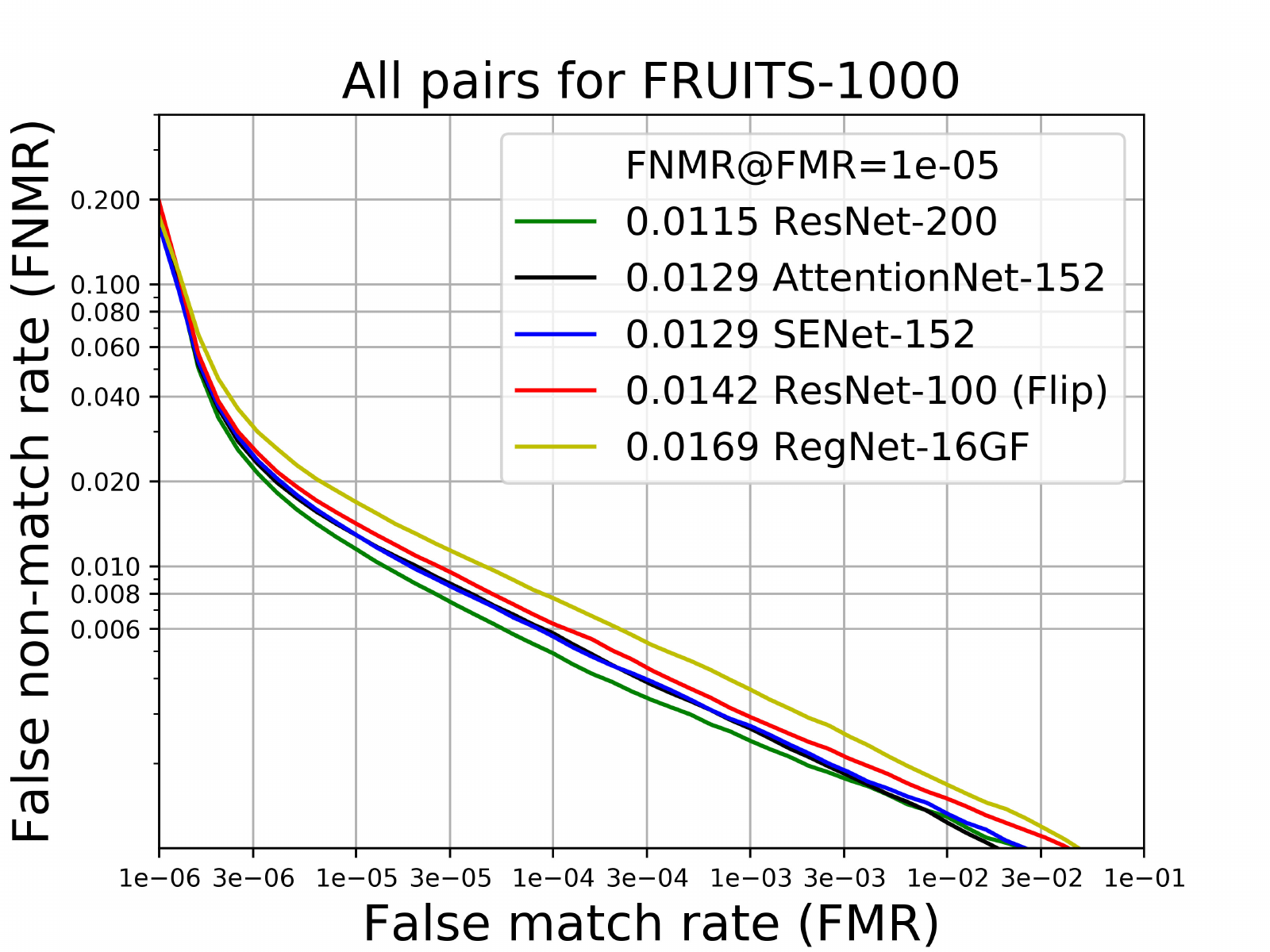}}
\subfigure[{\scriptsize Attributes for FRUITS-1000}]{
\label{fig:allattribute1000}
\includegraphics[width=0.43\linewidth]{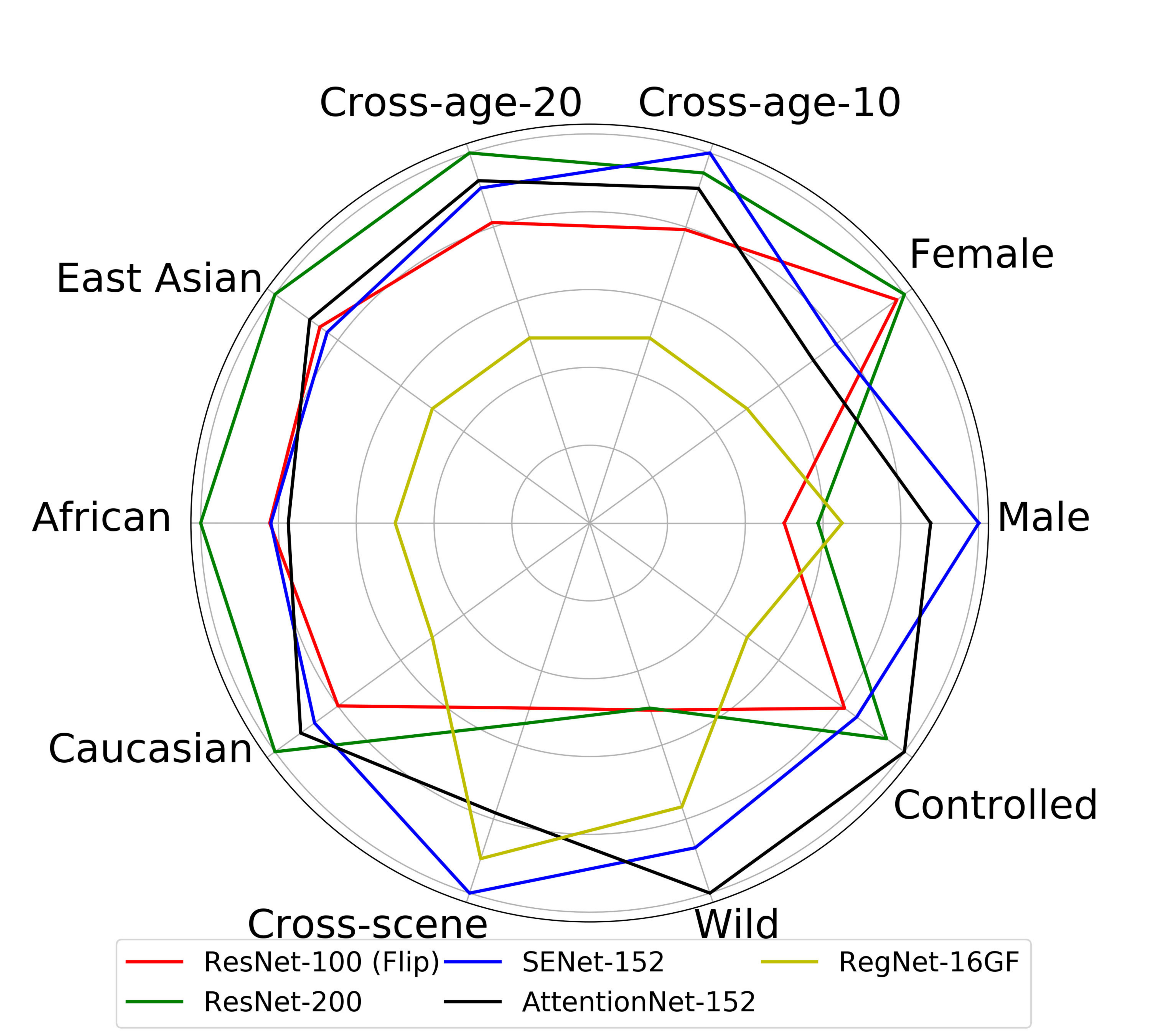}}
\vspace{-2mm}
\caption{Comprehensive performance comparisons of different models under the proposed FRUITS protocols. Left part is the FMR-FNMR plot for \emph{All} pairs verification, and models are ranked in legend according to FNMR@FMR=1e-5 (\textbf{lower FNMR is better}). The right part shows the attribute plots under FNMR@FMR=1e-5, which is normalized to 0.5-1.0 for better visualization (\textbf{outer is better}).}
\vspace{-1mm}
\label{fig:fruits}
\end{figure}

Due to strict time limitation, models constrained by FRUITS-100 can only adopt lightweight architectures, including RetinaFace-MobileNet-0.25 \cite{RetinaFace} for face detection and alignment, ResNet-14, MobileFaceNet (Flip), EfficientNet-B0 and RegNet-800MF for face feature extraction. FNMR on \emph{All} pairs and analysis of attribute bias are shown in Fig.\ref{fig:allfmr100} and Fig.\ref{fig:allattribute100}. Because of the weak detection and recognition modules, the best baseline (RegNet-800MF) only obtains 5.88\% FNMR@FMR=1e-5 (lower is better). Therefore, there leaves a substantial room for future improvement under the FRUITS-100 protocol.

For the FRUITS-500 protocol, we can employ more capable modern networks, such as RetinaFace-ResNet-50 \cite{RetinaFace} for pre-processing, and ResNet-100, ResNet-50 (Flip), SENet-50, ResNeXt-100, RegNet-8GF for feature embedding. As shown in Fig.\ref{fig:allfmr500} and Fig.\ref{fig:allattribute500}, ResNet-100 exhibits best overall performance in unbiased face verification. ResNet-50 with flip testing achieves lowest FNMR according to the attribute indicators of \emph{Wild} and \emph{Male}, while ResNeXt ranks first in the \emph{Cross-scene} track.

Recognition models under the FRUITS-1000 protocol can be more complicated and powerful, therefore we explore ResNet-100 (Flip), ResNet-200, SENet-152, AttentionNet-152 and RegNet-16GF for face feature embedding. As shown in Fig.\ref{fig:allfmr1000} and Fig.\ref{fig:allattribute1000}, ResNet-200 performs best in face verification and wins five attribute comparisons, while SENet-152 and AttentionNet-152 achieve three and two first-place respectively, according to the attribute indicator. Compared with lightweight FRUITS-100 track, performance of different large models are much closer. This result implies that new designs need to be explored for heavyweight FRUITS track.

\begin{table}[t]
\centering
\begin{center}{\scalebox{0.68}{
\smallskip
\begin{tabular}{c|c|c|c|c|c}
\hline

Protocol & Det\&Align & Embedding   & FLOPs & Params & Time \\

\hline\hline
\multirow{4}{*}[-1.2ex]{\tabincell{c}{FRUITS\\-100}}
& M-0.25    & ResNet-14 & 2.1G & 19.2M& 97ms \\
& M-0.25    & MobileFaceNet (Flip) &230.3M &1.2M & 65ms \\
& M-0.25    & EfficientNet-B0 &394.2M &11.6M& 94ms\\
& M-0.25    & RegNet-800MF &831.0M &23.4M& 89ms \\ \hline

\multirow{4}{*}[-1.2ex]{\tabincell{c}{FRUITS\\-500}}
& R-50    & ResNet-100  &12.1G &65.2M & 481ms \\
& R-50    & ResNet-50 (Flip)  &6.3G & 43.6M& 492ms  \\
& R-50    & SENet-50  &6.3G & 43.8M&  374ms \\
& R-50    & ResNeXt-100  &8.2G &56.2M & 411ms  \\
& R-50    & RegNet-8GF  & 8.0G&82.7M& 429ms \\ \hline

\multirow{5}{*}[-1.2ex]{\tabincell{c}{FRUITS\\-1000}}
& R-50    & ResNet-100 (Flip)  & 12.1G&65.2M & 826ms \\
& R-50    & ResNet-200  &23.9G &109.3M & 892ms \\
& R-50    & SENet-152  & 18.1G& 101.0M& 792ms \\
& R-50    & AttentionNet-152  & 14.8G&61.3M & 785ms \\
& R-50    & RegNet-16GF &16.0G &103.7M&  772ms \\ \hline

\hline
\end{tabular}}}
\end{center}
\vspace{-4mm}
\caption{Settings and inference time of baselines. Loose cropped test images are resized to $224 \times 224$ for joint detection and alignment. M-0.25 and R-50 refer to RetinaFace using MobileNet-0.25 (23ms) and ResNet-50 (272ms) as the backbones. FLOPs and Params mean computational complexity and parameter number of recognition module, respectively. Time refers to the duration of the whole system.}
\label{table:time}
\vspace{-4mm}
\end{table}

\subsection{Results on NIST-FRVT}
Finally, we report the submission to the NIST-FRVT. Following the settings of FRUITS-1000, our system is built based on RetinaFace-ResNet-50 for detection and alignment, and ArcFace-ResNet-200 trained on WebFace42M for feature embedding. The inference is accelerated by OpenVINO~\cite{openvino} and the flip test is adopted. The final inference time is near 1300 milliseconds according to the NIST-FRVT report, meeting the latest 1500 milliseconds limitation. Tab.\ref{table:nist} illustrates top-ranking entries measured by FNMR across five tracks. Our model trained on the WebFace42M achieves overall 3rd among 430 submissions, showing impressive performance across different tracks. Considering hundreds of company entries to NIST-FRVT, the WebFace42M takes a significant step towards closing the data gap between academia and industry.

\begin{table}[!t]
\begin{center}{\scalebox{0.75}{
\begin{tabular}{c|c|c|c|c|c|c}
\hline
Rank & entries & Visa & Mugshot & VisaBorder  & Border  & Wild\\ \hline
\hline
1 & deepglint & 0.0027 & 0.0033 & 0.0043 & 0.0084 &  0.0301 \\
2 & visionlabs &0.0025  & 0.0029 & 0.0035 &  0.0064&  0.0306 \\ \hline
3 & ours & 0.0034 & 0.0028 & 0.0046  & 0.0088 &  0.0303 \\ \hline
4 & dahua & 0.0046 & 0.0049 & 0.0046 & 0.0076 &  0.0300 \\
5 & cib & 0.0061 & 0.0041 & 0.0048 & 0.0578 &  0.0302 \\  \hline
\end{tabular}}}
\end{center}
\vspace{-0.4cm}
\caption{Results on NIST-FRVT. Our Arcface model using ResNet-200 is trained on WebFace42M. FNMR at corresponding FMR is reported.}
\label{table:nist}
\vspace{-0.6cm}
\end{table}

\section{Discussion and Conclusion}
\vspace{-0.1cm}

\noindent{\bf Discussion} WebFace260M may exist bias in different attributes, which has been considered in some aspects. First, our initial name list is constructed from Freebase and IMDB, which contains great diversity. Second, the bias is also considered in the test set construction, metrics and baselines results . In the test set construction, our multi-ethnic annotators are encouraged to gather attribute-balanced faces. In experiments, we evaluate models over these attributes (\eg race, gender and age) and show the relative ranks. For real-world applications, existing of bias may cause performance drop over certain attributes.
Considering the extremely large faces in our WebFace260M, we can sample balanced data to train models with less bias.
Besides, recent de-bias face recognition researches \cite{RFW,wang2019mitigate,gong2020mitigating,gong2020jointly} may also alleviate this problem to some extent.
For the ethics of gathering dataset, detailed rules are listed in our website. In summary, we will provide strict access for applicants who sign license, and try our best to guarantee it for research purposes only.

\noindent{\bf Conclusion} In this paper, we dive into million-scale face recognition, contributing a high-quality training data with 42M images of 2M identities by using automatic cleaning, a test set containing rich attributes, a time-constrained evaluation protocol, a distributed framework at linear acceleration, a succession of baselines, and a final SOTA model. Equipped with this face benchmark, our model significantly reduces 40\% failure rate on the IJB-C dataset and ranks the 3rd among 430 entries on NIST-FRVT. We hope this benchmark could facilitate future research of large-scale face recognition.

{\small
\bibliographystyle{ieee_fullname}
\bibliography{egbib}
}

\end{document}